\documentclass[conference]{IEEEtran}
\IEEEoverridecommandlockouts
\usepackage{cite}
\usepackage{amsmath,amssymb,amsfonts}
\usepackage{algorithmic}
\usepackage{array}
\usepackage{graphicx}
\usepackage{textcomp}
\usepackage{multirow}
\usepackage{xcolor}
\def\BibTeX{{\rm B\kern-.05em{\sc i\kern-.025em b}\kern-.08em
    T\kern-.1667em\lower.7ex\hbox{E}\kern-.125emX}}
\begin{document}

\title{Guided Co-Attention with Adaptive Gating for Multimodal Medical Data Fusion \\
}

\author{\IEEEauthorblockN{Muntaqim Ahmed Raju\IEEEauthorrefmark{1}, Priyanka Siddappa\IEEEauthorrefmark{1}, Md Shifat Haider Al Amin\IEEEauthorrefmark{1}\\
Ruizhe Ma\IEEEauthorrefmark{1}, and Md Ripon Ahammed\IEEEauthorrefmark{2}}
\IEEEauthorblockA{\IEEEauthorrefmark{1}University of Massachusetts Lowell, USA}
\IEEEauthorblockA{\IEEEauthorrefmark{2}NYC Health \& Hospitals/Jacobi-NCB, USA}
\IEEEauthorblockA{Email: \{MuntaqimAhmed\_Raju, Priyanka\_Siddappa, MdShifat\_AlAmin\}@student.uml.edu\\
Ruizhe\_Ma@uml.edu, Ripon.Ahammed90@gmail.com}
}

\maketitle

\begin{abstract}
Multimodal fusion requires not only exchanging information across modalities
but also determining which representations should participate in the
interaction and how complementary cross-modal views should be combined. We
introduce Guided Co-Attention with Adaptive Gating, a fusion framework that
integrates token-wise guidance, bidirectional cross-attention, and
sample-specific fusion. The proposed method constructs two directional
representations: an image-conditioned text representation, obtained by
allowing textual tokens to query visual features, and a text-conditioned
image representation formed in the reverse direction. Token-wise guidance
modulates the representations before cross-attention, while an adaptive gate
learns the relative contribution of the two resulting directional
representations for each sample. We evaluate the approach using paired chest
radiographs and radiology reports as an image--text test bed for binary
(normal vs. abnormal) classification. The two cross-attention directions
exhibit distinct performance characteristics, demonstrating that reversing
the query and source modalities does not produce interchangeable
representations. Ablation experiments further show that bidirectional
interaction, adaptive gating, and token guidance provide complementary
benefits. The proposed model achieves 96.05\% accuracy, 96.10\% F1-score,
98.37\% AUC, 95.60\% Sens@95Spec, and 96.51\% Spec@95Sens,
yielding the highest mean performance across all evaluated metrics. These findings support explicitly modeling interaction direction, token relevance, and sample-dependent integration when constructing
multimodal representations.
\end{abstract}

\begin{IEEEkeywords}
Multimodal fusion, guided co-attention, bidirectional cross-modal interaction, adaptive gating, image--text representation learning, chest X-ray classification.
\end{IEEEkeywords}

\section{Introduction}

Multimodal fusion requires learning interactions between heterogeneous
representations~\cite{baltruvsaitis2018multimodal,raju2025multimodal}.
In image--text learning, visual representations preserve spatial structure,
whereas language representations encode contextual and semantic relationships.
Effective fusion therefore requires more than mapping both modalities into a
shared feature space: it must determine how information from one modality
modifies the representation of the other and how the resulting cross-modal
information should be integrated. Attention provides a natural mechanism for modeling such interactions.
Bilinear attention explicitly captures pairwise relationships between visual
and linguistic features~\cite{kim2018bilinear}, while Transformer-based
multimodal architectures exchange information through cross-attention
between modality-specific streams~\cite{lu2019vilbert,tan2019lxmert}.
Cross-modal attention is inherently directional because the query modality
determines the representation being updated by information retrieved from
the source modality~\cite{tsai2019multimodal}. Consequently, reversing the
query and source roles produces a different cross-conditioned
representation. Bidirectional interaction therefore provides two views of the same multimodal
sample. In image--text learning, textual queries can retrieve relevant visual
information, whereas visual queries can incorporate contextual textual
information~\cite{nguyen2018improved}. However, simply averaging the two
directions assumes that their relative contribution should remain fixed
across samples. Because the usefulness of the two cross-conditioned
representations may vary with the input, a more flexible fusion mechanism
should preserve both directional views before determining how they contribute
to the final representation. The quality of cross-modal interaction also depends on the token
representations supplied to attention. Although attention determines
pairwise relevance between query and source tokens, all valid tokens initially
participate in the interaction. In heterogeneous multimodal sequences,
individual tokens can differ substantially in task relevance
~\cite{lu2016hierarchical,yu2019deep}. Our previous work, CheXFuse~\cite{raju2026chexfuse}, addressed this issue by combining token-wise guidance with differential joint attention to promote selective multimodal interactions. While CheXFuse applied guidance within a shared image–text representation, the role of token guidance in explicitly directional cross-modal interaction and adaptive integration remains to be investigated.

Motivated by these observations, we propose Guided Co-Attention with
Adaptive Gating. The method applies token-wise guidance before
cross-attention, performs interaction in both cross-modal directions, and
preserves the resulting directional representations until the final fusion
stage. A learned gate conditioned jointly on both representations then
produces a sample-dependent interpolation between them. The distinction from conventional co-attention lies not simply in performing cross-attention in both directions, but in explicitly separating three stages of fusion: token relevance is regulated before interaction, directional
cross-conditioned representations are preserved independently, and their
relative contribution is determined only after cross-modal interaction.
Guidance and gating therefore act at different levels of the fusion process. We evaluate this formulation using paired chest radiographs and radiology
reports as an image--text test bed. A radiograph encodes spatial patterns
associated with anatomy and pathology, while the corresponding report
describes radiographic observations using contextual clinical language.
Prior medical vision--language methods, including GLoRIA, ConVIRT and BioViL, have
demonstrated useful global and local image--text correspondences in this
setting~\cite{huang2021gloria,zhang2022convirt,boecking2022making}. These properties make
radiograph--report pairs a useful setting for examining directional
multimodal fusion.

The main contributions of this work are as follows:
\begin{itemize}
    \item We formulate multimodal fusion using two explicitly preserved
    cross-conditioned representations produced by opposite cross-attention
    directions.

    \item We incorporate token-wise guidance into both cross-attention directions to regulate the representations participating in each directional interaction.

    \item We introduce sample-conditioned adaptive gating for integrating the
    two directional representations and evaluate the contributions of
    bidirectional interaction, adaptive fusion, and token guidance through
    controlled ablations.
\end{itemize}

\section{Method}
\label{sec:method}

\begin{figure*}[t]
\centering
\includegraphics[width=\textwidth]{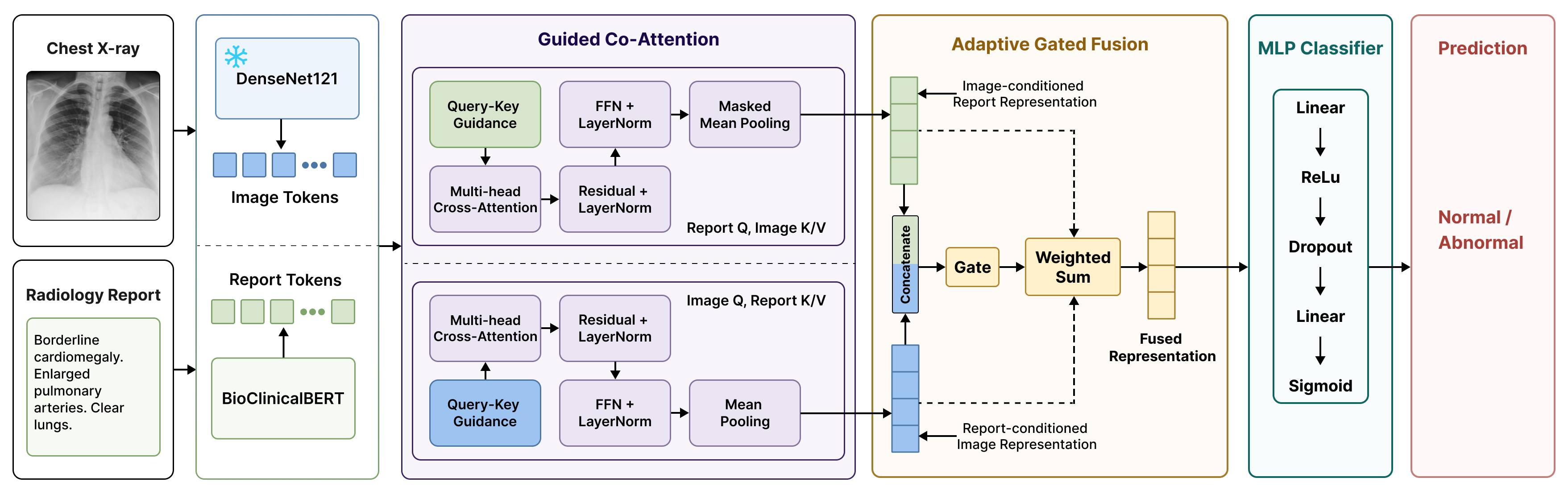}
\caption{Overview of the proposed Guided Co-Attention with Adaptive Gating
method, instantiated for paired chest radiograph--report fusion. Modality-specific
tokens interact through guided cross-attention in two complementary directions.
The resulting cross-conditioned representations are pooled and combined through
a sample-specific adaptive gate for downstream prediction.}
\label{fig:framework}
\end{figure*}

\subsection{Overview and Problem Formulation}
\label{subsec:overview}

We consider multimodal fusion between two tokenized representations,
$\mathbf{X}^{A}\in\mathbb{R}^{N_A\times d}$ and
$\mathbf{X}^{B}\in\mathbb{R}^{N_B\times d}$, where $N_A$ and $N_B$
denote the numbers of tokens in the two modalities and $d$ is their shared
embedding dimension. The objective is to construct a fused representation
that captures information exchanged in both cross-modal directions while
allowing the relative contribution of each directional interaction to vary
across samples.

The proposed method consists of three stages. First, token-wise guidance
modulates the representations entering cross-attention. Second, guided
cross-attention is performed in both directions, producing two complementary
cross-conditioned representations. Third, a sample-specific adaptive gate
determines the relative contribution of the two directional representations
to the final fused feature.

The fusion formulation itself is independent of a particular modality pair.
For evaluation in this work, we instantiate the two modalities using a frontal
chest radiograph and its corresponding radiology report. Given an image--report
pair $(\mathbf{x}^{I},x^{T},y)$, where $\mathbf{x}^{I}$ denotes the radiograph,
$x^{T}$ denotes the report, and $y\in\{0,1\}$ is the downstream class label,
the image and report are first encoded as spatial visual tokens and contextual
textual tokens, respectively. Figure~\ref{fig:framework} illustrates this
experimental instantiation.

\subsection{Modality-Specific Token Representations}
\label{subsec:token_representations}

\subsubsection{Visual Representation}

For the chest radiograph modality, we use a DenseNet121 encoder initialized
with CheXNet-14 pretrained weights~\cite{rajpurkar2017chexnet}. The convolutional backbone produces a
$7\times7$ feature map with 1024 channels. ReLU activation is applied to the
feature map, followed by a learnable $1\times1$ convolution that projects
each spatial location to the common multimodal embedding dimension $d=768$.

The projected feature map is flattened over its spatial dimensions to obtain
$N_I=49$ visual tokens. Layer normalization and a learnable positional
embedding are subsequently applied, yielding the image-token representation

\begin{equation}
\mathbf{X}^{I}
\in
\mathbb{R}^{N_I\times d}.
\end{equation}

Each token therefore corresponds to a spatial location in the projected
radiographic feature map while retaining positional information required for
cross-modal interaction.

\subsubsection{Textual Representation}

The corresponding radiology report is tokenized and encoded using
BioClinicalBERT~\cite{alsentzer2019publicly}. Each report is padded or truncated to a maximum sequence
length of $N_T=128$. An attention mask

\begin{equation}
\mathbf{m}^{T}\in\{0,1\}^{N_T}
\end{equation}

identifies valid report tokens and distinguishes them from padding positions.
The final hidden states of the language encoder form the contextual textual
representation

\begin{equation}
\mathbf{X}^{T}
\in
\mathbb{R}^{N_T\times d}.
\end{equation}

Because BioClinicalBERT produces $d=768$ dimensional hidden states, the textual
tokens already match the multimodal embedding dimension and are supplied to
the fusion module without an additional projection.

\subsection{Guided Cross-Attention}
\label{subsec:gca}

Standard cross-attention estimates relationships between a query sequence and
a source sequence from their token representations. In the proposed method,
token-wise guidance is introduced before this interaction so that the
representations participating in cross-attention can be modulated according
to their learned relevance.

Let
$\mathbf{Q}^{(0)}\in\mathbb{R}^{N_Q\times d}$,
$\mathbf{K}^{(0)}\in\mathbb{R}^{N_K\times d}$, and
$\mathbf{V}^{(0)}\in\mathbb{R}^{N_K\times d}$
denote the original query, key, and value sequences. Separate lightweight
guidance networks estimate a scalar coefficient for each query and source
token:

\begin{equation}
\begin{aligned}
\boldsymbol{\alpha}^{Q}
&=
\sigma\left(
\mathcal{D}\left[
\delta\left(
\mathbf{Q}^{(0)}\mathbf{W}^{Q}_{1}
+\mathbf{b}^{Q}_{1}
\right)
\right]
\mathbf{W}^{Q}_{2}
+b^{Q}_{2}
\right),\\
\boldsymbol{\alpha}^{K}
&=
\sigma\left(
\mathcal{D}\left[
\delta\left(
\mathbf{K}^{(0)}\mathbf{W}^{K}_{1}
+\mathbf{b}^{K}_{1}
\right)
\right]
\mathbf{W}^{K}_{2}
+b^{K}_{2}
\right),
\end{aligned}
\label{eq:guidance}
\end{equation}

where $\delta(\cdot)$ denotes ReLU activation,
$\sigma(\cdot)$ denotes the sigmoid function, and
$\mathcal{D}(\cdot)$ denotes dropout. The resulting guidance vectors

\begin{equation}
\boldsymbol{\alpha}^{Q}\in\mathbb{R}^{N_Q\times1},
\qquad
\boldsymbol{\alpha}^{K}\in\mathbb{R}^{N_K\times1}
\end{equation}

provide an independently learned modulation coefficient for each query and
source token.

The guided query, key, and value representations are then defined as

\begin{equation}
\widetilde{\mathbf{Q}}
=
\mathbf{Q}^{(0)}
\odot
\boldsymbol{\alpha}^{Q},
\qquad
\widetilde{\mathbf{K}}
=
\mathbf{K}^{(0)}
\odot
\boldsymbol{\alpha}^{K},
\qquad
\widetilde{\mathbf{V}}
=
\mathbf{V}^{(0)}
\odot
\boldsymbol{\alpha}^{K},
\label{eq:guided_tokens}
\end{equation}

where $\odot$ denotes element-wise multiplication with broadcasting over the
feature dimension. The same source-token guidance is applied to the keys and
values so that learned token relevance influences both attention-score
estimation and subsequent feature aggregation.

Multi-head cross-attention is then applied to the guided representations:

\begin{equation}
\mathbf{C}
=
\operatorname{MHA}
\left(
\widetilde{\mathbf{Q}},
\widetilde{\mathbf{K}},
\widetilde{\mathbf{V}};
\mathbf{M}^{K}
\right),
\label{eq:cross_attention}
\end{equation}

where $\mathbf{M}^{K}$ denotes the key-padding mask associated with
the source sequence. The mask is used when textual tokens serve as keys and
values so that padded report positions do not participate in the attention
operation. No source mask is required when the visual tokens provide the keys
and values because all spatial positions are valid.

The cross-attention output is combined with the original, unguided query
representation through a residual connection:

\begin{equation}
\begin{aligned}
\mathbf{H}^{(1)}
&=
\operatorname{LN}
\left(
\mathbf{Q}^{(0)}
+
\mathbf{C}
\right),\\
\mathbf{H}^{(2)}
&=
\operatorname{LN}
\left(
\mathbf{H}^{(1)}
+
\operatorname{FFN}
\left(
\mathbf{H}^{(1)}
\right)
\right),
\end{aligned}
\label{eq:gca_output}
\end{equation}

where $\operatorname{LN}(\cdot)$ denotes layer normalization. Retaining the
original query sequence in the residual pathway allows token guidance to
regulate the representations participating in cross-modal interaction without
replacing the original query representation itself.

The feed-forward network expands the feature dimension from $d$ to $4d$,
applies GELU activation and dropout, and projects the representation back to
$d$. We denote the resulting output of a guided cross-attention block as

\begin{equation}
\mathbf{O}
=
\operatorname{GCA}
\left(
\mathbf{Q}^{(0)},
\mathbf{K}^{(0)},
\mathbf{V}^{(0)};
\mathbf{M}^{K}
\right)
=
\mathbf{H}^{(2)}.
\label{eq:gca_definition}
\end{equation}

\subsection{Bidirectional Cross-Modal Interaction}
\label{subsec:bidirectional}

A single cross-attention direction conditions one modality on information
retrieved from the other. Because reversing the query and source roles
produces a different cross-conditioned representation, we apply guided
cross-attention in both directions.

\subsubsection{Image-Conditioned Text Representation}

In the first direction, textual tokens act as queries, while visual tokens
provide the keys and values:

\begin{equation}
\mathbf{O}^{T\leftarrow I}
=
\operatorname{GCA}_{T\leftarrow I}
\left(
\mathbf{X}^{T},
\mathbf{X}^{I},
\mathbf{X}^{I}
\right).
\label{eq:text_image_branch}
\end{equation}

The resulting sequence

\begin{equation}
\mathbf{O}^{T\leftarrow I}
\in
\mathbb{R}^{N_T\times d}
\end{equation}

preserves the textual token structure while incorporating visual information
retrieved from the spatial image representation.

Because the report sequence contains padding positions, masked mean pooling
is applied:

\begin{equation}
\mathbf{z}^{T\leftarrow I}
=
\frac{
\sum_{j=1}^{N_T}
m^{T}_{j}
\mathbf{O}^{T\leftarrow I}_{j,:}
}{
\sum_{j=1}^{N_T}
m^{T}_{j}
}.
\label{eq:text_pooling}
\end{equation}

\subsubsection{Text-Conditioned Image Representation}

In the reverse direction, visual tokens act as queries, while textual tokens
provide the keys and values:

\begin{equation}
\mathbf{O}^{I\leftarrow T}
=
\operatorname{GCA}_{I\leftarrow T}
\left(
\mathbf{X}^{I},
\mathbf{X}^{T},
\mathbf{X}^{T};
\mathbf{M}^{T}
\right),
\label{eq:image_text_branch}
\end{equation}

where $\mathbf{M}^{T}$ is the key-padding mask constructed from
$\mathbf{m}^{T}$. This prevents padded textual positions from contributing
to cross-attention.

The output

\begin{equation}
\mathbf{O}^{I\leftarrow T}
\in
\mathbb{R}^{N_I\times d}
\end{equation}

preserves the spatial visual-token structure while incorporating contextual
information retrieved from the report representation. Since every image token
corresponds to a valid spatial location, mean pooling is applied:

\begin{equation}
\mathbf{z}^{I\leftarrow T}
=
\frac{1}{N_I}
\sum_{i=1}^{N_I}
\mathbf{O}^{I\leftarrow T}_{i,:}.
\label{eq:image_pooling}
\end{equation}

The two directional representations therefore provide complementary
cross-conditioned views of the multimodal sample. The representation
$\mathbf{z}^{T\leftarrow I}$ retains the textual perspective while
incorporating retrieved visual information, whereas
$\mathbf{z}^{I\leftarrow T}$ retains the visual perspective while
incorporating retrieved textual context.

\subsection{Adaptive Gated Fusion}
\label{subsec:adaptive_fusion}

Although the two directional representations encode information from the same
multimodal pair, their relative usefulness need not be identical for every
sample. We therefore avoid combining them through a fixed averaging operation
and instead learn a sample-specific fusion coefficient.

The two pooled directional representations are first concatenated, and the
resulting $2d$-dimensional feature is supplied to a two-layer gating network:

\begin{equation}
g
=
\sigma\left(
\left(\mathbf{w}^{g}_{2}\right)^{\top}
\mathcal{D}\left[
\delta\left(
\mathbf{W}^{g}_{1}
\left(
\mathbf{z}^{T\leftarrow I}
\Vert
\mathbf{z}^{I\leftarrow T}
\right)
+
\mathbf{b}^{g}_{1}
\right)
\right]
+
b^{g}_{2}
\right),
\label{eq:fusion_gate}
\end{equation}

where $\Vert$ denotes feature concatenation and $g\in[0,1]$. The first
gating layer maps the concatenated $2d$-dimensional feature to $d$
dimensions, followed by ReLU activation and dropout. The second layer
produces a single scalar coefficient.

The final multimodal representation is obtained as

\begin{equation}
\mathbf{z}^{F}
=
g\,\mathbf{z}^{T\leftarrow I}
+
(1-g)\,\mathbf{z}^{I\leftarrow T}.
\label{eq:gated_fusion}
\end{equation}

Because $g$ is conditioned jointly on both directional representations, the
fusion rule can vary from one multimodal sample to another. Equation
\eqref{eq:gated_fusion} therefore performs a sample-specific convex
combination of the image-conditioned textual representation and the
text-conditioned visual representation rather than imposing equal
contributions across all samples.

\subsection{Downstream Prediction}
\label{subsec:prediction}

The fusion method produces $\mathbf{z}^{F}\in\mathbb{R}^{d}$ as its final
multimodal representation. For the binary classification task used in our
evaluation, this representation is passed through a two-layer classification
head:

\begin{equation}
\mathbf{h}^{C}
=
\mathcal{D}
\left[
\delta
\left(
\mathbf{W}^{C}_{1}\mathbf{z}^{F}
+
\mathbf{b}^{C}_{1}
\right)
\right],
\label{eq:classifier_hidden}
\end{equation}

followed by

\begin{equation}
\ell
=
\left(\mathbf{w}^{C}_{2}\right)^{\top}
\mathbf{h}^{C}
+
b^{C}_{2},
\qquad
\hat{y}
=
\sigma(\ell),
\label{eq:classifier_output}
\end{equation}

where $\ell$ is the classification logit and $\hat{y}$ denotes the predicted
probability of the positive class. The classification head is used only for
the downstream evaluation; the multimodal fusion mechanism itself is defined
by the guided bidirectional interaction and adaptive gated combination
described above.

\section{Experimental Evaluation}
\label{sec:experiments}

\subsection{Data}
\label{subsec:data}

We use the Indiana University (IU) Chest X-ray dataset~\cite{demner2016preparing} as the image--text test bed for evaluating the proposed multimodal fusion method. Each sample in the processed dataset consists of one frontal chest radiograph, its corresponding original radiology report, and a binary (normal vs. abnormal) label. The original report text, stored in the \texttt{Findings} field, is used as the textual input, while the binary (normal vs. abnormal) label is derived from the \textit {MeSH major} field. Abnormal samples are assigned label 1, whereas normal samples are assigned label 0. For image feature extraction, the DenseNet121 encoder is initialized using the CheXNet-14 pretrained weights introduced by Rajpurkar et al.~\cite{rajpurkar2017chexnet}. Since the pretrained image encoder was optimized using frontal-view radiographs, we retain only frontal images from the IU Chest X-ray dataset to maintain consistency between the pretraining and downstream experimental settings. After removing non-frontal images and samples without a valid image--report correspondence, the processed dataset contains 2,129 paired samples. The dataset is partitioned at the patient level into approximately 70\%, 20\%, and 10\% for training, validation, and testing, respectively. This results in 1,490 training samples, 428 validation samples, and 211 test samples. The training and validation sets are class-balanced, while the test set contains 106 abnormal and 105 normal samples. 

\subsection{Preprocessing}
\label{subsec:preprocessing}

For the visual modality, all chest radiographs are converted to three-channel
RGB images. During training, random resized cropping is applied to obtain
$224\times224$ inputs, followed by tensor conversion and normalization using
the ImageNet mean and standard deviation. During validation and testing,
images are resized to $256\times256$ and center-cropped to $224\times224$
before normalization.
For the textual modality, the \textit{Findings} section of each report is
tokenized using the BioClinicalBERT tokenizer. Reports are padded or
truncated to a maximum sequence length of 128 tokens. The corresponding
attention mask is retained so that padded positions can be excluded during
cross-attention and pooling.

\subsection{Training Setup}
\label{subsec:training}

All experiments are implemented in PyTorch. For the multimodal models, the
DenseNet121 image encoder is initialized with CheXNet-14 pretrained weights
and kept frozen during training. The visual projection layer, multimodal
fusion components, classification head, and BioClinicalBERT encoder are
optimized jointly. BioClinicalBERT is therefore fine-tuned during multimodal
training.

All multimodal representations use a shared embedding dimension of
$d=768$. The cross-attention modules use 8 attention heads, and a dropout
rate of 0.2 is applied within the fusion and classification components. The
classification head consists of a fully connected layer followed by ReLU,
dropout, and a final linear layer that produces a single logit for binary
classification.

The multimodal models are trained using binary cross-entropy with logits
loss and optimized with AdamW using an initial learning rate of
$1\times10^{-4}$ and weight decay of $1\times10^{-4}$. A StepLR scheduler
is applied with a step size of 3 and decay factor $\gamma=0.1$. We use a
batch size of 16 and train for a maximum of 10 epochs.

Model selection is based on validation accuracy. Early stopping is applied
with a patience of three epochs and a minimum improvement threshold of
$1\times10^{-4}$. For each run, the checkpoint with the highest validation
accuracy is retained for evaluation on the held-out test set. The same
encoder configuration, data partition, and optimization protocol are used
across all multimodal variants so that performance differences primarily
reflect the fusion mechanism being evaluated.

For the image-only baseline, the CheXNet-14 pretrained DenseNet121 backbone
is fine-tuned jointly with the binary classification head. This baseline is
optimized using AdamW with a learning rate of $1\times10^{-3}$ and weight
decay of $1\times10^{-4}$, while retaining the same batch size,
learning-rate scheduler, maximum number of epochs, and early-stopping
protocol used for the multimodal experiments.

Each experiment is repeated using three random seeds: 7, 42, and 123. Final
results are reported as mean $\pm$ standard deviation across the three runs.
\subsection{Baselines and Fusion Variants}
\label{subsec:baselines}

We evaluate unimodal references and reduced multimodal configurations to
separate the contributions of directional interaction, bidirectional fusion,
adaptive gating, and token-wise guidance. All multimodal configurations use
the same image and text encoders, embedding dimension, data partitions, and
training protocol, and differ only in the fusion components being evaluated.

\subsubsection{Image-only}
The visual baseline uses the CheXNet-pretrained DenseNet121 representation
followed by a binary classification head.

\subsubsection{Text-only}
The textual baseline uses pooled BioClinicalBERT representations from the
radiology \textit{Findings} section followed by binary classification.

\subsubsection{Image-conditioned text}
This configuration retains only the $T\leftarrow I$ direction, in which
textual tokens query visual tokens. The pooled output therefore preserves a
textual representation conditioned on retrieved visual information.

\subsubsection{Text-conditioned image}
This configuration retains only the $I\leftarrow T$ direction, in which
visual tokens query textual tokens. The pooled output preserves a visual
representation conditioned on retrieved textual information.

\subsubsection{Ungated bidirectional fusion}
Both directional cross-attention pathways are retained, while adaptive
gating and token guidance are omitted. Their pooled representations are
combined using fixed equal-weight averaging,
$\mathbf{z}^{F}_{\mathrm{mean}}
=0.5(\mathbf{z}^{T\leftarrow I}+\mathbf{z}^{I\leftarrow T})$.
This configuration provides a fixed-fusion reference for evaluating the
benefit of adaptive directional weighting.

\subsubsection{Gated bidirectional fusion}
Both directional pathways are retained, but fixed averaging is replaced by
the proposed sample-conditioned gate. Token-wise guidance is disabled.
Comparison with the ungated bidirectional configuration therefore isolates
the effect of adaptive fusion.

\subsubsection{Guided Co-Attention with Adaptive Gating}
The complete model applies token-wise guidance before both cross-attention
directions and combines the resulting representations using adaptive gating.
Comparison with the gated bidirectional configuration isolates the additional
effect of token-wise guidance.

\subsection{Evaluation Metrics}
\label{subsec:metrics}

Performance is evaluated using accuracy, F1-score, area under the receiver
operating characteristic curve (AUC), sensitivity at 95\% specificity
(Sens@95Spec), and specificity at 95\% sensitivity (Spec@95Sens). The
abnormal class is treated as the positive class throughout the evaluation.
Accuracy and F1-score are computed using a fixed decision threshold of 0.5,
while AUC measures discrimination across the full range of decision
thresholds. To additionally evaluate performance under high-specificity and
high-sensitivity operating conditions, we report Sens@95Spec and
Spec@95Sens.
For Sens@95Spec, the decision threshold is selected on the validation set as
the operating point whose specificity is closest to 95\%. This threshold is
then applied unchanged to the held-out test set, and the resulting test
sensitivity is reported. Similarly, for Spec@95Sens, the threshold is selected
on the validation set as the operating point whose sensitivity is closest to
95\%, and the corresponding test specificity is reported using the same
threshold.
All metrics are computed independently for each random seed and are reported
as mean $\pm$ standard deviation over the three runs.

\section{Results and Analysis}
\label{sec:results}
\subsection{Unimodal and Multimodal Reference Performance}
\label{subsec:reference}

\begin{table*}[!t]
\caption{Performance comparison between unimodal baselines, our previous CheXFuse framework, and the proposed multimodal fusion model. Results are reported as mean $\pm$ standard deviation in percentage points over three random seeds (7, 42, and 123).}
\label{tab:main_results}
\centering
\begin{tabular}{lccccccc}
\hline
Model
& Image
& Text
& Accuracy
& F1-score
& AUC
& Sens@95Spec
& Spec@95Sens \\
\hline

Image-only
& \checkmark
& --
& $70.46 \pm 0.72$
& $68.76 \pm 2.12$
& $75.88 \pm 0.67$
& $18.24 \pm 3.03$
& $29.21 \pm 1.10$ \\

Text-only
& --
& \checkmark
& $95.26 \pm 0.07$
& $95.31 \pm 0.05$
& $97.89 \pm 0.15$
& $93.40 \pm 1.89$
& $95.24 \pm 0.95$ \\

CheXFuse~\cite{raju2026chexfuse}
& \checkmark
& \checkmark
& $95.73 \pm 0.04$
& $95.79 \pm 0.02$
& --
& $95.28 \pm 0.94$
& -- \\

\textbf{Proposed}
& \checkmark
& \checkmark
& $\mathbf{96.05 \pm 0.99}$
& $\mathbf{96.10 \pm 0.98}$
& $\mathbf{98.37 \pm 0.59}$
& $\mathbf{95.60 \pm 2.37}$
& $\mathbf{96.51 \pm 0.55}$ \\
\hline
\end{tabular}
\end{table*}

We first establish reference performance for the individual modalities and
the proposed multimodal model. The unimodal results characterize the
predictive information available from each modality independently, while the
proposed model measures the performance obtained when bidirectional
cross-modal interaction, adaptive gating, and token-wise guidance are used
jointly. Table~\ref{tab:main_results} reports the results over three
independent runs using seeds 7, 42, and 123.

The two modalities exhibit substantially different standalone performance.
The image-only model achieves $70.46\%$ accuracy, $68.76\%$ F1-score, and
$75.88\%$ AUC. Its Sens@95Spec and Spec@95Sens are $18.24\%$ and
$29.21\%$, respectively, indicating considerably weaker class separation
under constrained operating points. In contrast, the text-only model provides a strong unimodal reference, achieving $95.26\%$ accuracy, $95.31\%$ F1-score, and $97.89\%$ AUC, together with $93.40\%$ Sens@95Spec and $95.24\%$ Spec@95Sens. This strong performance is expected in part from the nature of the textual input used in our evaluation. Specifically, the textual modality is derived from the \textit{Findings} section of the radiology report, which contains the radiologist's description of observations present in the corresponding radiograph. Because these descriptions can directly characterize normal findings as well as radiographic abnormalities, the resulting textual
representation is closely related to the (normal vs. abnormal) downstream
label and consequently contains highly discriminative semantic information.
The text-only model therefore constitutes a particularly demanding unimodal
reference for evaluating whether multimodal fusion can contribute useful
information beyond the interpreted radiographic findings.

The proposed multimodal model achieves the highest mean performance across
all evaluated metrics, reaching $96.05\%$ accuracy, $96.10\%$ F1-score,
and $98.37\%$ AUC. Relative to the text-only reference, these results
correspond to absolute improvements of $0.79$ percentage points in accuracy,
$0.79$ percentage points in F1-score, and $0.48$ percentage points in AUC.
The improvement is more pronounced at the constrained operating points.
Sens@95Spec increases from $93.40\%$ to $95.60\%$, corresponding to a
$2.20$ percentage-point gain, while Spec@95Sens increases from $95.24\%$
to $96.51\%$, corresponding to a $1.27$ percentage-point gain. Although the overall gains are modest given the strength of the textual representation, their consistency across all five metrics suggests that the visual modality provides complementary information not fully captured by the radiologist's descriptions.

Compared with our previous CheXFuse framework~\cite{raju2026chexfuse}, the proposed method improves mean accuracy from $95.73\%$ to $96.05\%$,
F1-score from $95.79\%$ to $96.10\%$, and Sens@95Spec
from $95.28\%$ to $95.60\%$. These modest numerical
improvements suggest that explicitly preserving complementary
cross-modal representations and adaptively integrating their
contributions may provide additional benefits over joint
attention-based fusion.

To investigate the contributions of individual architectural components, we next examine the two cross-attention directions separately, followed by the effects of bidirectional fusion, adaptive gating, and token-wise guidance.

\subsection{Directional Cross-Modal Interaction}
\label{subsec:directional}

\begin{table*}[!t]
\caption{Component-wise ablation of the proposed multimodal fusion method.
Results are reported as mean $\pm$ standard deviation in percentage points
over three random seeds (7, 42, and 123). Best and second-best mean results
are shown in bold and underlined, respectively.}
\label{tab:ablation_results}
\centering

\begin{tabular}{lccccc}
\hline
Configuration
& Accuracy
& F1
& AUC
& Sens@95Spec
& Spec@95Sens \\
\hline

Image-conditioned text
& $95.10 \pm 1.52$
& $95.08 \pm 1.64$
& $97.47 \pm 0.51$
& $\underline{94.65 \pm 1.44}$
& $\underline{95.87 \pm 0.55}$ \\

Text-conditioned image
& $95.58 \pm 0.27$
& $95.64 \pm 0.31$
& $96.35 \pm 0.32$
& $69.81 \pm 14.83$
& $94.60 \pm 0.55$ \\

Ungated bidirectional
& $95.58 \pm 0.55$
& $95.66 \pm 0.58$
& $97.61 \pm 0.93$
& $83.33 \pm 7.63$
& $94.92 \pm 0.55$ \\

Bidirectional + gate
& $\underline{95.73 \pm 0.47}$
& $\underline{95.80 \pm 0.29}$
& $\underline{97.66 \pm 0.63}$
& $93.08 \pm 4.65$
& $95.56 \pm 1.10$ \\

\textbf{Proposed}
& $\mathbf{96.05 \pm 0.99}$
& $\mathbf{96.10 \pm 0.98}$
& $\mathbf{98.37 \pm 0.59}$
& $\mathbf{95.60 \pm 2.37}$
& $\mathbf{96.51 \pm 0.55}$ \\
\hline

\end{tabular}
\end{table*}

Table~\ref{tab:ablation_results} first examines the two cross-attention
directions independently. The image-conditioned text branch
($T\leftarrow I$), in which textual tokens query visual tokens, achieves
$95.10\%$ accuracy, $95.08\%$ F1-score, and $97.47\%$ AUC. It also
achieves $94.65\%$ Sens@95Spec and $95.87\%$ Spec@95Sens.

The text-conditioned image branch ($I\leftarrow T$), in which visual tokens
query textual tokens, achieves slightly higher accuracy and F1-score of
$95.58\%$ and $95.64\%$, respectively. However, its AUC decreases to
$96.35\%$, while Sens@95Spec decreases substantially to
$69.81\% \pm 14.83\%$. The comparatively large variation in Sens@95Spec
also indicates less stable behavior across runs under the high-specificity
operating constraint.

The two directions therefore exhibit distinct performance characteristics.
The text-conditioned image representation provides slightly stronger
default-threshold accuracy and F1-score, whereas the image-conditioned text
representation provides stronger AUC and substantially better constrained
operating-point performance. These differences indicate that reversing the
query and source modalities does not produce interchangeable
representations and motivate retaining both directional views during
multimodal fusion.

\subsection{Effect of Bidirectional Fusion}
\label{subsec:bidirectional_effect}

We next examine whether retaining both cross-attention directions provides
an advantage over using either direction independently. The ungated
bidirectional configuration incorporates both
$\mathbf{z}^{T\leftarrow I}$ and $\mathbf{z}^{I\leftarrow T}$ while
excluding the adaptive gating and token-guidance mechanisms.

The ungated bidirectional model achieves $95.58\%$ accuracy, $95.66\%$
F1-score, and $97.61\%$ AUC. Relative to the text-conditioned image branch,
bidirectional interaction preserves the same mean accuracy, increases
F1-score by $0.02$ percentage points, and improves AUC by $1.26$ percentage
points. Sens@95Spec also increases from $69.81\%$ to $83.33\%$, a gain of
$13.52$ percentage points.

Compared with the image-conditioned text branch, however, the ungated
bidirectional model improves accuracy and F1-score by $0.48$ and $0.58$
percentage points, respectively, but does not preserve its stronger
constrained operating-point behavior. Sens@95Spec decreases from $94.65\%$
to $83.33\%$, while Spec@95Sens decreases from $95.87\%$ to $94.92\%$.

These results show that simply retaining both interaction directions does
not uniformly inherit the strongest characteristics of each directional
representation. Although bidirectional interaction improves several overall
metrics, the relative contribution of the two directions remains important,
particularly under constrained operating conditions. This observation
motivates the use of an adaptive mechanism for balancing the two
cross-conditioned representations.

\subsection{Effect of Adaptive Gating}
\label{subsec:gating_effect}

To isolate the contribution of adaptive gating, we compare the ungated and
gated bidirectional configurations. Both models retain the same two
cross-attention directions, while the gated variant additionally learns a
sample-specific coefficient for combining their pooled representations. Introducing adaptive gating increases accuracy from $95.58\%$ to $95.73\%$
and F1-score from $95.66\%$ to $95.80\%$, corresponding to gains of $0.15$
and $0.14$ percentage points, respectively. AUC increases slightly from
$97.61\%$ to $97.66\%$. A larger effect is observed at the constrained operating points. Sens@95Spec increases from $83.33\%$ to $93.08\%$, corresponding to a
$9.75$ percentage-point improvement, while Spec@95Sens increases from
$94.92\%$ to $95.56\%$, a gain of $0.64$ percentage points. The standard
deviation of Sens@95Spec also decreases from $7.63$ to $4.65$ percentage
points across the three runs. The relatively modest improvements in accuracy, F1-score, and AUC, together with the substantially larger gain in Sens@95Spec, suggest that adaptive gating is particularly beneficial when the two directional representations must be balanced under a constrained operating condition. Rather than
assigning the same relative contribution to both directions for every
sample, the learned gate allows their influence on the fused representation
to vary according to the input pair.

\subsection{Effect of Token Guidance}
\label{subsec:guidance_effect}

Finally, we isolate the contribution of token-wise guidance by comparing
the proposed model with the unguided gated bidirectional configuration.
Both variants use the same two cross-attention directions and the same
adaptive gating mechanism; the only difference is whether query and source
tokens are modulated before cross-attention.
Introducing token guidance increases accuracy from $95.73\%$ to $96.05\%$
and F1-score from $95.80\%$ to $96.10\%$, corresponding to gains of $0.32$
and $0.30$ percentage points, respectively. AUC increases from $97.66\%$
to $98.37\%$, yielding a $0.71$ percentage-point improvement.
The gains are also evident at the constrained operating points.
Sens@95Spec increases from $93.08\%$ to $95.60\%$, corresponding to a
$2.52$ percentage-point improvement, while Spec@95Sens increases from
$95.56\%$ to $96.51\%$, a gain of $0.95$ percentage points.
The proposed model therefore achieves the highest mean performance across
all five evaluation metrics. Since the gated bidirectional and proposed
configurations differ only in the use of token guidance, this comparison
more directly isolates the contribution of pre-attention token modulation.
The improvements in AUC and constrained operating-point performance are
consistent with guidance providing more selective representations for the
subsequent cross-modal interaction.

Taken together, the ablation results reveal a progressive contribution from
the proposed fusion components. The two cross-attention directions encode
distinct information, their ungated combination provides a joint
bidirectional representation, adaptive gating improves the balance between
the two directions, and token-wise guidance provides a further improvement
when applied before cross-modal interaction.

\section{Discussion}
\label{sec:discussion}

The ablation results show that cross-attention direction materially affects
the resulting multimodal representation. The image-conditioned text branch
provides stronger AUC and constrained operating-point performance, whereas
the text-conditioned image branch achieves slightly higher accuracy and
F1-score. This indicates that the two directions capture different
cross-conditioned information rather than interchangeable representations.
One possible explanation is that using textual tokens as queries allows the
semantic structure of the report to constrain information retrieval from the
spatial visual representation, which may contribute to the stronger
discrimination observed for this direction. In the reverse direction, visual
tokens query an already highly informative textual representation, producing
strong default-threshold performance but less stable behavior under the
high-specificity constraint. The larger variability in Sens@95Spec for this
branch further highlights the different behavior of the two interaction
directions.
Combining the two directions through fixed equal-weight averaging does not
preserve the strongest characteristics of either branch. Although ungated
bidirectional fusion improves several overall metrics, its constrained
operating-point performance remains below that of the image-conditioned text
branch. This suggests that complementary directional representations are not
necessarily optimally integrated by assigning them equal importance for every
input. Preserving both views is therefore useful, but their relative
contribution also matters.
Adaptive gating provides a mechanism for relaxing this fixed-weight
assumption. Its largest improvement is observed in Sens@95Spec, whereas the
changes in accuracy, F1-score, and AUC are comparatively modest. This pattern
suggests that adaptive integration is particularly useful under the more
restrictive high-specificity operating condition. A plausible explanation is
that the relative usefulness of the two cross-conditioned representations is
not constant across samples. Because the gate is conditioned jointly on both
representations, the model has the capacity to adjust their relative
contribution rather than enforcing the same mixture for every input.
Token guidance provides an additional benefit after bidirectional interaction
and adaptive gating are already present. The improvement is more pronounced
in AUC and constrained operating-point metrics than in accuracy and F1-score.
This is consistent with pre-attention modulation improving the representations
from which cross-modal relationships are formed. Unlike adaptive gating, which
operates after the directional representations have been constructed,
guidance acts before cross-attention and regulates the token representations
participating in the interaction. The two mechanisms therefore address
different stages of the fusion process, which is consistent with the
additional improvement obtained when guidance is introduced on top of gated
bidirectional fusion.

Collectively, the findings support the central design of the proposed
framework: the direction of cross-modal information flow influences the
resulting representation, fixed integration does not fully exploit the two
directional views, and additional selectivity can be introduced both before
cross-attention through token guidance and after directional interaction
through adaptive gating.

\section{Limitations and Future Work}
\label{sec:limitations}

This study has several limitations. First, the proposed fusion mechanism is
evaluated on a single image--text dataset. Although the formulation itself is
not specific to radiology, experiments with additional datasets and modality
pairs are required to determine whether the observed benefits of directional
interaction, adaptive gating, and token guidance generalize beyond the current
setting. Second, the downstream evaluation is limited to binary (normal vs. abnormal)
classification. This provides a controlled setting for studying the fusion
mechanism, but it does not capture the complexity of multi-label disease
recognition or other multimodal prediction tasks. Future work will therefore
evaluate the method on larger datasets and more challenging downstream tasks,
including multi-label classification. Finally, the textual input is derived from the radiology \textit{Findings} section, which already contains a clinical interpretation of the corresponding image. This explains the strong predictive value of the textual modality and also limits how broadly the current results can be interpreted. Evaluation
with other forms of paired text, including less directly diagnostic
descriptions, would provide a stronger test of whether the proposed fusion
mechanism remains effective when complementary information is distributed more
evenly across modalities.

\section{Conclusion}
\label{sec:conclusion}

This work introduced Guided Co-Attention with Adaptive Gating, a multimodal
fusion framework that combines token-wise guidance, bidirectional
cross-attention, and sample-specific integration of directional
representations. The experimental findings show that the two cross-attention
directions are not interchangeable: each captures a different
cross-conditioned view of the paired modalities. Retaining both directions
provides complementary information, while adaptive gating improves how these
representations are combined and token guidance further refines the
information entering cross-modal interaction. The findings support the broader view that effective multimodal fusion should account not only for which modalities are
combined, but also for the direction of their interaction, the relevance of
the participating tokens, and the relative contribution of the resulting
cross-modal representations. Future work will examine the generality of these
findings across additional datasets, modality combinations, and more complex
prediction tasks.

\bibliographystyle{IEEEtran}
\bibliography{references}

\end{document}